\documentclass{darwin}

\usepackage{color}
\usepackage{amsmath,graphicx,array}
\usepackage{dcolumn,soul}%

\usepackage{amsthm}
\usepackage[figuresright]{rotating}%
\usepackage{algorithm, algorithmicx, algpseudocode}
\usepackage{listings}%
\usepackage{hyperref}
\usepackage{ragged2e}

\makeatletter
\def\uns{\ifmmode\,\else$\,$\fi}%

\usepackage{subfig}
\usepackage{multirow}
\usepackage{booktabs}
\makeatother

\jvol{XX}
\jnum{X}
\jyear{Year}
\doi{XXX/XXX/XXXX}
\received{XX XX Year}
\revised{XX XX Year}
\accepted{XX XX Year}

\begin{document}
	
%\dhead{RESEARCH ARTICLE}
	
%\subhead{PHYSICS}
	
\title{Darwin3: A large-scale neuromorphic chip with a Novel ISA and On-Chip Learning}
	
\author{De Ma$^{1,2,3,4 \dagger}$}
	
\author{Xiaofei Jin$^{1, 2 \dagger}$}
	
\author{Shichun Sun$^{2}$}
	
\author{Yitao Li$^{1,3}$}
	
\author{Xundong Wu$^{2}$}
	
\author{Youneng Hu$^{1}$}
	
\author{Fangchao Yang$^{2}$}

\author{Huajin Tang$^{1,2,3,4}$}
	
\author{Xiaolei Zhu$^{5,2}$}

\author{Peng Lin$^{1,3,4}$}
	
\author{Gang Pan$^{1,2,3,4*}$}

\affil{$^1$College of Computer Science and Technology, Zhejiang University, Hangzhou 310058, China}
	
\affil{$^2$Research Center for Intelligent Computing Hardware, Zhejiang Lab, Hangzhou 311121, China}

\affil{$^3$The State Key Lab of Brain-Machine Intelligence, Zhejiang University, Hangzhou 310027, China}

\affil{$^4$MOE Frontier Science Center for Brain Science and Brain-machine Integration, Zhejiang University, Hangzhou 310027, China}

\affil{$^5$College of Micro-Nano College of Micro-Nano Electronics, Zhejiang University, Hangzhou 311200, China}
	
\authornote{\textbf{Corresponding authors.} Email: gpan@zju.edu.cn}
\authornote{Equally contributed to this work.}
	
\abstract[ABSTRACT]{
Spiking Neural Networks (SNNs) are gaining increasing attention for their biological plausibility and potential for improved computational efficiency. To match the high spatial-temporal dynamics in SNNs, neuromorphic chips are highly desired to execute SNNs in hardware-based neuron and synapse circuits directly. This paper presents a large-scale neuromorphic chip named Darwin3 with a novel instruction set architecture(ISA), which comprises 10 primary instructions and a few extended instructions. It supports flexible neuron model programming and local learning rule designs. The Darwin3 chip architecture is designed in a mesh of computing nodes with an innovative routing algorithm. We used a compression mechanism to represent synaptic connections, significantly reducing memory usage. The Darwin3 chip supports up to 2.35 million neurons, making it the largest of its kind in neuron scale. The experimental results showed that code density was improved up to 28.3x in Darwin3, and neuron core fan-in and fan-out were improved up to 4096x and 3072x by connection compression compared to the physical memory depth. Our Darwin3 chip also provided memory saving between 6.8X and 200.8X when mapping convolutional spiking neural networks (CSNN) onto the chip, demonstrating state-of-the-art performance in accuracy and latency compared to other neuromorphic chips.}

%\jelcode{Pa, J6, P16, E22}
	
	
\maketitle
	
\section{INTRODUCTION}
Spiking neural networks (SNNs) have garnered significant attention from researchers due to their ability to process spatial-temporal information in an efficient event-driven manner. To exploit the capabilities of SNNs, several spiking neural network simulation platforms have been introduced, such as Brian2\cite{Brian2}, NEST\cite{NEST}, and SPAIC\cite{SPAIC}. 
Nevertheless, the dependence of these platforms on using extensive GPU and CPU resources to mimic the spiking dynamics with a high count of timing steps potentially diminish the intrinsic advantages of SNNs. Neuromorphic chips are designed for efficient execution of spiking neural networks, which have demonstrated promising performance in brain simulation and specific ultra-low power scenarios. However, several limitations prevent them from fully leveraging the advantages of spiking neural networks. To better leverage the benefits of the SNN models, we should emphasize the three aspects when designing neuromorphic chips:

\textbf{Flexibility of Neural Models:} One of the key functions of neuromorphic chips is to simulate diverse biological neurons and synapses. However, many neuromorphic chips only support a single type of neuron model, as evidenced in platforms like Neurogrid\cite{Neurogrid}, which is based on analog neuron circuits. Some works introduce a degree of configurability to accommodate various neuron models. Loihi\cite{Loihi} achieved enhanced learning capabilities through configurable sets of traces and delays. FlexLearn\cite{FlexLearn} has conceived a versatile data path that amalgamates key features from diverse models. Moreover, endeavors have been undertaken to develop fully configurable neuronal models using instructions. SpiNNaker's multi-core processors\cite{SpiNNaker}, based on conventional ARM cores, provide significant flexibility. However, it is associated with reduced performance and energy efficiency compared to other accelerators. While Loihi2\cite{Loihi2} presents an instruction set incorporating logical and mathematical operations similar to RISC instructions. However, instruction sets designed for conventional neural networks lack efficiency for SNNs despite their flexibility. 
	
\textbf{Synapse Density:} To further unlock the potential of SNNs, neuromorphic chips need to support the representation of large-scale SNNs with more complex topologies\cite{mapping_large}. However, current neuromorphic chips pay less attention to this aspect, primarily concentrating on simulating the behavior of neurons and synapses. For instance, TrueNorth\cite{TrueNorth} employs a crossbar design for synaptic connections, but it suffers from limited and fixed fan-in/fan-out capacity. Loihi~\cite{Loihi} takes an approach by using axon indexes to encode topology, thereby enhancing flexibility. Loihi2~\cite{Loihi2} proposes optimizing for convolutional and factorized connections but gives less attention to other connection types. Unicorn~\cite{Unicorn} introduces a technique for merging synapses from multiple cores to extend the synaptic scale of a single core. Thus, improving synapse density for various topologies under limited storage conditions is crucial for optimizing the cost-effectiveness of the chip.

\textbf{On-chip Learning Ability:} Learning capability is a critical feature of biological neural networks. Currently, only a few neuromorphic chips support on-chip learning. Among those, the supported learning rules are pretty restricted. For instance, BrainscaleS2\cite{BrainScaleS2} only accommodates fixed learning algorithms. Loihi\cite{Loihi} supports programmable rules for pre-, post-, and reward traces. Loihi2\cite{Loihi2} extends its capabilities of the programmable rules applied to pre-, post-, and generalized "third-factor" traces. However, even with the enhanced flexibility exhibited by Loihi2\cite{Loihi2}, it cannot accommodate novel learning rules that might emerge. The latest research achievements in the field of electrochemical memory array\cite{open_loop} also provide new reference solutions.
	
In this paper, we design a large-scale neuromorphic chip with a domain-specific Instruction Set Architecture (ISA), named Darwin3, to support model flexibility, system scalability, and on-chip learning capability of the chip. Darwin3 is the third generation of our Darwin \cite{darwin} family of neuromorphic chips, which was successfully taped out and lit up in December 2022. Our main contributions are as follows.

\textbf{1)} We propose a domain-specific instruction set architecture (ISA) for neuromorphic systems, capable of efficiently describing diverse models and learning rules, including the integrate-and-fire (LIF) family\cite{lifreview}, Izhikevich\cite{izhi_simple},  and STDP\cite{STDP}, among others. The proposed architecture excels in achieving high parallelism during computational operations, including loading parameters and updating state variables such as membrane potential and weights.

\textbf{2)} We design a novel mechanism to represent the topology of SNNs. This mechanism effectively compresses the information required to describe synaptic connections, thereby reducing overall memory usage.

The article is organized as follows: Firstly, we introduce the topic and briefly overview the article's contents. Second, we present the neuromorphic computing domain-specific ISA. Then, we offer the overall architecture of the neuromorphic chip and the implementation of each part, including the architecture of neuron nodes and the mechanism of topology representation. Lastly, we demonstrate the experiment results.

\section{THE DARWIN3 DOMAIN-SPECIFIC ISA} 
\subsection{Model Abstraction of Neurons, Synapses, and Learning}
\label{sec_bg}
Many neuron models have been proposed in the field of computational neuroscience. The leaky integrate-and-fire (LIF) family\cite{smith}\cite{ermentrout}\cite{adaptive}\cite{lifreview}  is a group of spiking neuron models that can be described by one or two-dimensional differential equations and were widely implemented on hardware accelerators. These models have been developed for use in many real-world applications various applications. The Hodgkin-Huxley model\cite{hodgkin1}\cite{hodgkin2} is considered biologically plausible and accurately captures the intricacies of neuron behavior with four-dimensional differential equations that represent the transfer of ions across the neuron membrane. However, this model can cause very high computational costs. The Izhikevich model\cite{izhi_simple}, specifically designed to replicate bursting and spiking behaviors observed in the Hodgkin-Huxley model, is represented with two-dimensional differential equations.
 
All these neuron models are represented using systems of differential equations, with variations occurring only in the number of equations and the variables and parameters in each equation. The primary operators needed to solve them are the same. Therefore, it can be a practical approach to identify the common features shared by complex LIF models and utilize them to construct more complex models by introducing additional state variables and computation steps. We chose the Adaptive Leaky Integrate-and-Fire (AdLIF) model\cite{adaptive} as the baseline with relatively more variables and parameters. Mathematically, it can be expressed by Equation\ref{equ1}, which captures the dynamics of the model and its adaptation properties.
	\vspace{-2 mm}
	\begin{small}
		\begin{align}{} \label{equ1}
			{\tau _m}\frac{{d{v_m}}}{{dt}} =  - ({v_m} - {E_L}) - \frac{1}{g}({v_{adp}} - I) \nonumber\\
			{\tau _{adp}}\frac{{d{v_{adp}}}}{{dt}} = a({v_m} - {E_L}) - {v_{adp}} \\
			if({v_m} > {v_{th}}):{v_m} = {v_{0}},{v_{adp}} = {v_{adp}} + b \nonumber
		\end{align}
	\end{small}
	
Where $v_m$ is the membrane potential, $\tau_m$ is the membrane time constant, $E_L$ is the leak reversal potential, $g$ is synapse conductance, $v_{adp}$ is adaptation current, $\tau_{adp}$ is time constant of the adaptation current, $a$ is the sensitivity to the sub-threshold fluctuations of the membrane potential, $b$ is the increment of $v_{adp}$ produced by a spike,$v_0$ is the reset potential after spike and $I$ is synaptic spike current.

	%Electrical synapses are a phenomenological model of chemical synapses and the most widely used model of SNNs. 
Similar to the various designs of neuron models with different computational complexity, there are also multiple synapse models, such as the delta and alpha synapse models\cite{dynamic}. One of the complex and commonly used models is the conductance-based(COBA) dual exponential model\cite{based}\cite{dynamic}, as shown in Equation \ref{equ2}, which has a similar computational complexity to that of Equation \ref{equ1}. We chose this model as our representative synapse model.
	\vspace{-2 mm}
	\begin{small}
		\begin{align}{} \label{equ2}
			\frac{{dh}}{{dt}} = \frac{{ - h}}{{{\tau _{rise}}}} + \delta ({t_0} - t) \nonumber\\
			\frac{{dg}}{{dt}} = \frac{{ - g}}{{{\tau _{decay}}}} + h\\
			I = g({v_m} - {E_L}) \nonumber
		\end{align}
	\end{small}
	
Where $\delta$ is a spike at time $t_0$, $h$ is the gating variable of the ion channel, $g$ is synapse conductance, $\tau_{decay}$ is the time constant of the synaptic decay phase, $\tau_{rise}$ is the time constant of the synaptic rise phase, $I$ is synaptic spike current, $v_m$ is the membrane potential and $E_L$ is the leak reversal potential.

Synaptic plasticity\cite{plasticity}, the ability of synapses to change their strength, was first proposed as a mechanism of learning and memory by Donald Hebb\cite{hebb}. After that, numerous learning rules have been proposed ever since. The STDP rule\cite{STDP}, and their variants are the most widely used. One relatively complex variant considers triplet\cite{triplet} interactions and is reward-modulated\cite{rstdp}. We select this model as the baseline, and through the selection of different state variables and parameters, the same equation can describe most STDP and its variant rules. The rule can be expressed mathematically as Equation \ref{equ3}.
	\vspace{-2 mm}
	\begin{small}
		\begin{align}{}\label{equ3}
			{\tau _{pre_0}}\frac{{d{x_0}}}{{dt}} =  - {x_{0}} + {a_{pre_0}}\delta (t - {t_{pre_0}}) \nonumber\\
			{\tau _{post_0}}\frac{{d{y_0}}}{{dt}} =  - {y_{0}} + {a_{post_0}}\delta (t - {t_{post_0}}) \nonumber\\
			{\tau _{post_1}}\frac{{d{y_1}}}{{dt}} =  - {y_1} + {a_{pos{t_1}}}\delta (t - {t_{post_1}}) \nonumber\\
			{\tau _{rwd}}\frac{{dr}}{{dt}} =  - r + {a_{rwd}}\delta (t - {t_{rwd}}) \\
			dw(t) = {A_{pre}}r{x_0}(t)\delta (t - {t_{pos{t_0}}}) \nonumber\\
			- {A_{pos{t_0}}}r{y_0}\delta (t - {t_{pre}}) - {A_{pos{t_1}}}r{y_1}\delta (t - {t_{pre}})\nonumber
		\end{align}
	\end{small}
	
Where $x_0$ is the pre-synaptic spike trace, $y_0$ is the first post-synaptic spike trace, $y_1$ is the second post-synaptic spike trace, $r$ is the reward to modulate the synaptic traces, $\tau_{*}$ are time constants of $x_0$, $y_0$, $y_1$ and $r$.

Equations \ref{equ1} to \ref{equ3} described three representative models. To implement the models using digital circuits, we need to convert the differential equations to a discrete form. By applying the Euler method, Equations \ref{equ1} and \ref{equ2}, are converted to Equation \ref{equ4} and \ref{equ5}.
        \vspace{-2 mm}
 	\begin{small}
		\begin{align}{}\label{equ4}
		v(t + 1) = {p_0}v(t) + {p_1}I(t + 1) + {p_2}{v_{adp}}(t + 1) + {c_0} \nonumber \\
		{v_{adp}}(t + 1) = {p_3}{v_{adp}}(t) + {p_4}v(t) + {c_1} \\
		if(v(t + 1) > {v_{{\rm{th}}}}):  \left\{ \begin{array}{l} \nonumber
				v(t + 1) = {v_0}  \nonumber\\
				{v_{adp}}(t + 1) = {v_{adp}}(t + 1) + {c_2}  \nonumber
			\end{array} \right.
		\end{align}
	\end{small}
        \vspace{-3 mm}
	\begin{small}
		\begin{align}{}\label{equ5}
		h(t + 1) = {p_8}h(t) + {w_{ij}}{\rm{H}}[t - {t^s}] \nonumber\\
		I(t + 1) = {\rm{g}}(t + 1)v(t) + {p_7}g(t + 1) \\
		g(t + 1) = {p_5}g(t) + {p_6}h(t + 1) \nonumber
        \end{align}
	\end{small}
 
	Where $p_0$ to $p_7$ are fixed coefficient parameters, $c_0$ to $c_2$ are constants. Similarly, Equation \ref{equ3} can be converted to the form shown in Equation \ref{equ6}. 
	\vspace{-2 mm}
	\begin{small}
		\begin{align}{}\label{equ6}
			{x_0}(t + 1) = P_3^*x(t) + C_0^*{x_2}(t)  \nonumber \\
			{y_0}(t + 1) = P_4^*{y_0}(t) + C_1^*{y_2}(t) \nonumber\\
			{y_1}(t + 1) = P_5^*{y_1}(t) + C_2^*{y_2}(t) \nonumber\\
			{r_0}(t) = P_6^*{r_0}(t) + C_3^*{r_2}(t)\\
			w(t + 1) = w(t) + P_0^*{r_0}(t){x_0}(t){y_2}(t) \nonumber\\
			+ P_1^*{r_0}(t){y_0}(t){x_2}(t) + P_2^*{r_0}(t){y_1}(t){x_2}(t) \nonumber
		\end{align}
	\end{small}
	
	Where $P_0^*$ to $P_6^*$ are fixed coefficient parameters, $C_0^*$ to $C_3^*$ are constants.
 
\begin{figure}[h]
		\centering
		\subfloat[]{\includegraphics[scale=0.6]{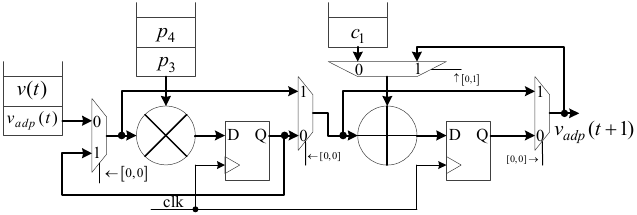}} \vspace{-5 mm}
		\hfil
		\subfloat[]{\includegraphics[scale=0.6]{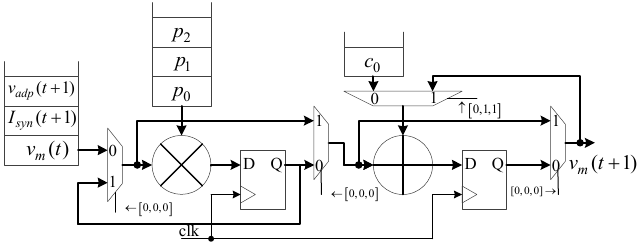}} \vspace{-5 mm}
		\hfil
		\subfloat[]{\includegraphics[scale=0.6]{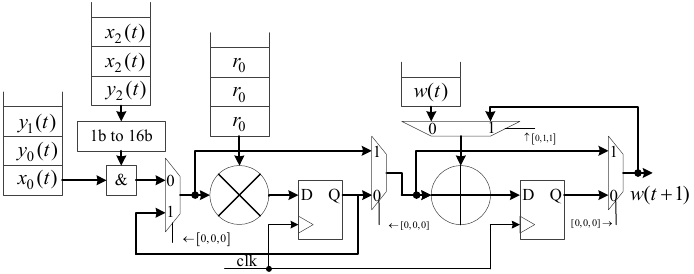}} \vspace{-5 mm}
		\hfil
		\subfloat[]{\includegraphics[scale=0.6]{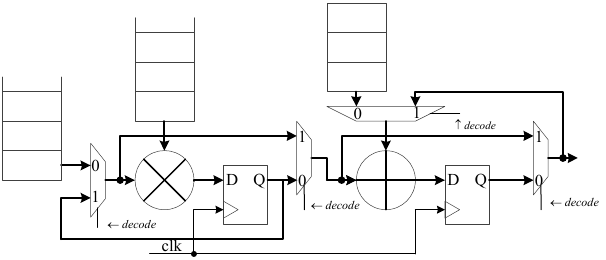}}
		\caption{Typical Data Path. (a) The Data Path of $v_{adp}$. (b) The Data Path of $v_{m}$. (c) The Data Path of $w$. (d) The Common Data Path for State Variables.}
		\label{fig_data_path}
	\end{figure}

Equations \ref{equ4}, \ref{equ5}, and \ref{equ6} reveal that both complex LIF models and STDP variants can be expressed as polynomials involving multiple multiplication and addition operations. To implement these polynomial computations in digital circuits, we map them to corresponding data paths for further analysis. Figure \ref{fig_data_path} (a) (b) and (c) illustrates that the data paths of $v_{adp}$, and $v_m$ in Equations \ref{equ4}, \ref{equ5} and $w$ in Equation \ref{equ6} are almost identical, except for different control signals from selectors and input sources, which allows us to efficiently implement these computations in circuits using a unified data path, where parameters can be pre-configured statically, and state variables are updated continuously over time steps. For more complex cases such as Izhikevich model\cite{izhi_simple}, the parameter that multiplies the state variables in the computation process is also a state variable. Therefore, we obtain the unified data path shown in Figure \ref{fig_data_path} (d).
	
	\begin{table*}[!t]
		\caption{Registers and Instruction Set}
		\label{regs}
		\centering
		(a) Registers for State Variables and Parameters \\
		\scalebox{0.763}{
			\begin{tabular}{|c|m{7cm}|c|m{7cm}|}
				\hline
				\multicolumn{1}{|c|}{\textbf{Name}} & \multicolumn{1}{c|}{\textbf{Description}} &\multicolumn{1}{c|}{\textbf{Name}} & \multicolumn{1}{c|}{\textbf{Description}} \\ \hline
				$v_{m}$, $S_{0}$   & The membrane potential  & $X_{0}$-$X_{1}$    & Two presynaptic spike traces $LS_{0}$-$LS_{1}$    \\ \hline
				g, $S_{1}$         & The synaptic conductance  & $X_{2}$            & Flag for spike from pre-synapse $L_{2}$    \\ \hline
				I, $S_{2}$         & The synaptic current     & $Y_{0}$-$Y_{1}$    & Two postsynaptic spike traces $LS_{3}$-$LS_{4}$  \\ \hline
				h, $S_{3}$         & The gating variable  &  $Y_{2}$            & Flag for spike from post-synapse $LS_{5}$   \\ \hline
				$v_{adp}$, $S_{4}$ & The adaptive voltage & $R_{0}$-$R_{1}$    & traces to do reward and punishment $LS_{6}$-$LS_{7}$  \\ \hline
				$v_{th}$, $S_{5}$  & The threshold voltage  & $R_{2}$            & Reward and Punishment $LS_{8}$   \\ \hline
				w                  & The synaptic weight  & $LP_{0}$-$LP_{7}$  & Eight parameters for learning stage, corresponding to $P^{*}$ in Equation \ref{equ6}\\ \hline
				$v_0$              & The reset membrane potential value & $LC_{0}$-$LC_{7}$  & Eight constants for learning stage, corresponding to $C^{*}$ in Equation \ref{equ6} \\ \hline
				$IP_{0}$-$IP_{7}$  & Eight parameters for inference stage, corresponding to $p_0$-$p_7$ in Equations \ref{equ4} and \ref{equ5} & $TR_{0}$-$TR_{7}$ & Temporary registers for parameters and states      \\ \hline
				$IC_{0}$-$IC_{2}$  & Three constants for inference stage, corresponding to $c_0$-$c_2$ in Equation \ref{equ4} and & \ref{equ5} - & \multicolumn{1}{c|}{-}\\ \hline		
			\end{tabular}
		}
		\\ \hfill
		
		(b) Neuromorphic Specific Instruction Set \\
		\scalebox{0.75}{
			%\begin{tabular}{|c|m{15.5cm}|}\hline
			%\begin{tabular}{|c|p{2cm}|p{3cm}|p{3cm}|p{3cm}|p{3cm}|}\hline
			\begin{tabular}{|c|c|c|c|c|c|}\hline
				\multicolumn{1}{|c|}{\textbf{Opcode (5 bits)}} & \multicolumn{5}{c|}{\textbf{Operand (11 bits)}} \\ \hline
				\multirow{2}{*}{LSIS} & LS (1-bit) & \multicolumn{1}{m{4cm}<{\centering}|}{Reserve} & \multicolumn{3}{m{8.5cm}<{\centering}|}{NHIS (6-bit)}  \\
				\cline{2-6}
				& \multicolumn{5}{m{15.5cm}|}{LS indicates whether an operation is a load or store operation, and NHIS is a 6-hot code corresponding to the state variables $S_0$-$S_5$, indicating whether each state variable needs to be loaded or stored during the operation.} \\ \hline

				\multirow{2}{*}{LDIP} & \multicolumn{3}{m{8.5cm}<{\centering}|}{NHIP(8-bit)} & \multicolumn{2}{c|}{NHIC(3-bit)} \\
				\cline{2-6}
				& \multicolumn{5}{m{15.5cm}|}{NHIP is an 8-hot code corresponding to the parameters $p_0$-$p_7$, and NHIC is a 3-hot code corresponding to $c_0$-$c_2$, indicating whether each needs to be loaded during the operation.} \\ \hline

				\multirow{2}{*}{LSLS} & LS(1-bit) & \multicolumn{4}{c|}{NHLS(10-bit)} \\
				\cline{2-6}
				& \multicolumn{5}{m{15.5cm}|}{A one hot code LS bit indicates whether an operation is a load or store operation, and a 10-hot code NHLS corresponding to the state variables $LS_0$ to $LS_9$ indicates whether each state variable needs to be loaded or stored during the operation.}  \\ \hline
				
				\multirow{2}{*}{LDLP} & \multicolumn{3}{c|}{NHLP(7-bit)} &\multicolumn{2}{c|}{NHLC(4-bit)}
				\\ 
				\cline{2-6}				
				& \multicolumn{5}{m{15.5cm}|}{A 7-hot code NHLP corresponding to the parameters $LP_0$-$LP_6$ and a 4-hot code NHLC corresponding to $LC_0$-$LC_3$, indicate whether each parameter needs to be loaded during the operation.} \\ \hline

				\multirow{2}{*}{UPTIS} & Reserve & \multicolumn{1}{m{4cm}<{\centering}|}{OHIS (3-bit)} & \multicolumn{3}{m{8.5cm}<{\centering}|}{NHIP (6-bit)}  \\
				\cline{2-6}
				& \multicolumn{5}{m{15.5cm}|}{OHIS is a one-hot code indicating whether $I$, $g$, or $v_{adp}$ to be calculated and NHIP is a 6-hot code corresponding to $p_3$-$p_7$ and $c_1$, indicating whether each needs to participate in the calculation according to Equation \ref{equ4} and \ref{equ5}.} \\ \hline
			
				\multirow{2}{*}{UPTVM   } & \multicolumn{3}{c|}{Reserve} & \multicolumn{2}{c|}{NHVM(4-bit)} \\
				\cline{2-6}
				& \multicolumn{5}{m{15.5cm}|}{NHVM is a 4-hot code to determine whether $v_m$, $I$, $v_{adp}$, or $c_0$, needs to participate in the calculation to update $v_m$ according to Equation \ref{equ4} and \ref{equ5}.}  \\ \hline

				\multirow{2}{*}{UPTLS} & \multicolumn{2}{c|}{k(3-bit)} & \multicolumn{1}{m{3cm}<{\centering}|}{l(3-bit)} & \multicolumn{1}{m{3cm}<{\centering}|}{m(3-bit)} & \multicolumn{1}{c|}{n(2-bit)} \\
				\cline{2-6}	
				& \multicolumn{5}{m{15.5cm}|}{The variables k, l, m, and n determine the selected state variable $LS_k$ that needs to update according to the equation: $LS_k$(t+1)=$LP_l$*$LS_m$(t)+$LC_n$.} \\ \hline
				
				\multirow{2}{*}{UPTWT} & \multicolumn{2}{c|}{m(2-bit)} & \multicolumn{3}{m{9cm}<{\centering}|}{n(9-bit)}
				\\
				\cline{2-6}	
				& \multicolumn{5}{m{15.5cm}|}{A binary code m and n-hot code n determine the synaptic weight to update according to the equation: WT(t+1)=WT(t)+$L{P_m}\prod\limits {L{S_n}}$.} \\ \hline

				\multirow{2}{*}{UPTTS} & \multicolumn{2}{m{2cm}<{\centering}|}{k(3-bit)} &  \multicolumn{1}{m{3cm}<{\centering}|}{l(3-bit)}	&  \multicolumn{1}{m{3cm}<{\centering}|}{m(3-bit)}	& \multicolumn{1}{c|}{ n(2-bit)} \\				
				\cline{2-6}
				& \multicolumn{5}{m{15.5cm}|}{The variables k, l, m, and n determine the selected temporary state variable $RT_k$ that needs to update according to the equation: $RT_k$(t+1)=$P_l$*$S_m$(t)+$C_n$.}  \\ \hline				
								
				\multirow{2}{*}{GSPRS} & \multicolumn{3}{c|}{Reserve} &	 \multicolumn{2}{c|}{NHSP(4-bit)} \\
				\cline{2-6}
				& \multicolumn{5}{m{15.5cm}|}{A 4-hot code NHSP respectively determines whether to fire a spike, whether to perform threshold comparison, whether to involve adaptive operation, and whether membrane potential needs to reset to $v_0$.}  \\ \hline 
																				
			\end{tabular}
		} \\ \hfill

		(c) Extended Instruction List \\
		\scalebox{0.75}{
		\begin{tabular}{|c|l|c|l|}
			\hline
			\textbf{Type} & \textbf{Instructions} & \textbf{Type} & \textbf{Instructions} \\ \hline
			Arithmetic    & ADD SUB MUL ADDI      & Jump          & CMP JMP               \\ \hline
			Bitwise       & SHIFT LOGIC           & Memory        & SA, TS, LOAD/PUSH STORE/POP SP   \\ \hline
			Move          & MOV WMOV              & Pseudo        & DIV EXP               \\ \hline
		\end{tabular}
		}
		
	\end{table*}

	\subsection{The Proposed Darwin3 ISA}
	\label{sec_isa}
	
To effectively manage the data path and maximize performance and concurrency, it is crucial to design an efficient controller. First, we map state variables and parameters into a set of registers, as indicated in Table \ref{regs}(a). It covers the state variables and parameters related to neurons and synapses, in which constants are static parameters. To provide users with the flexibility to implement different models, we propose a specialized instruction set architecture (ISA), as shown in Table \ref{regs} (b).

The core principle of this ISA is to amalgamate common operations into a single instruction, taking into account the computational characteristics of SNNs. By doing so, it not only reduces the memory usage required for instructions but also minimizes the time needed for instruction decoding during the computation process. We defined a set of instructions outlined in Table \ref{regs} (b). This instruction set comprises ten primary commands. The first group, which focuses on load and store operations, consists of LSIS, LDIP, LSLS, and LDLP. Specifically, LSIS and LSLS cater to loading or writing back state variables for both the inference and learning processes, executed in parallel. On the other hand, LDIP and LDLP are designated for loading parameters in parallel for inference and learning phases, respectively.
	
The second group, tailored for updating state variables, includes UPTIS, UPTVM, UPTLS, UPTWT, and UPTTS. Among these, UPTIS updates state variables, excluding the membrane potential. UPTVM is exclusively for adjusting the membrane potential. UPTLS emphasizes state variables specific to the learning stage, while UPTWT manages the adjustments of synaptic weights. UPTTS oversees the updating of temporary state variables. Lastly, the GSPRS instruction is dedicated to generating spikes. With these instructions, we can effectively manage the computing units and support the process required for constructing flexible SNN models. 
	
Models such as AdEx\cite{adaptive} and HH\cite{hodgkin1} necessitate intricate operations, including exponential and division functions. These are not directly supported by the instructions outlined in Table \ref{regs} (b). The design enables users to perform division and exponentiation operations using computational units like shift, multiplication, and lookup tables, thus conserving hardware resources. Consequently, we've augmented our instruction set with several instructions typically found in reduced instruction set architectures, as detailed in Table \ref{regs} (c).

In Table \ref{example_neuron}, we present a range of code examples beyond basic loading and storing operations to illustrate the efficacy of the proposed ISA. This selection demonstrates that simple LIF models and more complex Triplet STDP rules can be concisely represented using minimal instructions. Additionally, specialized rules such as SDSP\cite{SDSP}  can be efficiently encoded through a strategic combination of instructions. This versatility underscores the high flexibility and effectiveness of our instruction set design, establishing it as a viable tool for researchers and developers engaged in implementing diverse models in SNNs.
	
	\begin{table}[!t]		
		\centering
		\caption{Code Examples for Widely Used Models}
		\label{example_neuron}
		\scalebox{0.6}{
			\begin{tabular}{|c|l|c|l|}
				\hline
				\multicolumn{1}{c|}{\textbf{Model}} & \multicolumn{1}{c|}{\textbf{Code}} & \multicolumn{1}{c|}{\textbf{Model}} & \multicolumn{1}{c|}{\textbf{Code}} \\ \hline
				LIF  \begin{tabular}[c]{@{}l@{}} \cite{smith} \end{tabular}          &  \begin{tabular}[c]{@{}l@{}}UPTVM 0xD\\ GSPRS 0xA\end{tabular} & \multicolumn{1}{l|}{Basic STDP \begin{tabular}[c]{@{}l@{}}  \cite{STDP} \end{tabular}} & \begin{tabular}[c]{@{}l@{}}UPTLS 0x20\\ UPTLS 0x08\\ UPTWT 0x108\\ UPTWT 0x260\end{tabular} \\ \hline
				QIF  \begin{tabular}[c]{@{}l@{}}  \cite{lifreview} \end{tabular}     & \begin{tabular}[c]{@{}l@{}}UPTTS   0x021\\ MOV      P0, RT0\\ UPTVM  0xD\\ GSPRS  0xA\end{tabular} & \multicolumn{1}{l|}{Triplet STDP  \begin{tabular}[c]{@{}l@{}}   \cite{triplet} \end{tabular} } &  \begin{tabular}[c]{@{}l@{}}UPTLS 0x20\\ UPTLS 0x08\\ UPTLS 0x04\\ UPTWT 0x10C\\ UPTWT 0x264\\ UPTWT 0x454\end{tabular} \\ \hline
				ExpIF   \begin{tabular}[c]{@{}l@{}}  \cite{lifreview} \end{tabular}       & \begin{tabular}[c]{@{}l@{}}UPTTS    0x061\\ EXP RT1  RT0\\ UPTTS    0x48A\\ MOV C0   RT2\\ UPTVM   0xD\\ GSPRS   0xA\end{tabular} & RSTDP  \begin{tabular}[c]{@{}l@{}} \cite{rstdp} \end{tabular}  & \begin{tabular}[c]{@{}l@{}}UPTLS 0x20\\ UPTLS 0x08\\ UPTLS 0x02\\ UPTWT 0x10C\\ UPTWT 0x264\end{tabular} \\ \hline
				\multicolumn{1}{|l|}{Izhikevich  \begin{tabular}[c]{@{}l@{}}  \cite{izhi_simple} \end{tabular}} & \begin{tabular}[c]{@{}l@{}}UPTTS   0x143\\ MOV      P0, RT0\\ UPTIS   0x38\\ UPTVM 0x0F\\ GSPRS   0xE\end{tabular} & S-TP  \begin{tabular}[c]{@{}l@{}} \cite{ANP} \end{tabular} &  \begin{tabular}[c]{@{}l@{}}LayerH:\\ \ \ UPTTS 0x15A \\ \ \ MOV LP0 RT0 \\ \ \ UPTWT 0xC0\\LayerO:\\ \ \ UPTWT 0x228\end{tabular} \\ \hline
				\multicolumn{1}{|c|}{-}  & \multicolumn{1}{c|}{-} & SDSP\begin{tabular}[c]{@{}l@{}}  \cite{SDSP} \end{tabular}  & \begin{tabular}[c]{@{}l@{}}UPTLS 0x20 \\ CMP  RT0 S0 \\ JMP  Keep \\ CMP  RT1 LS1 \\ JMP  Keep \\ CMP  LS1 RT3 \\ JMP Keep \\ CMP  LS1 RT2 \\ JMP  UP \\ SUB  W RT4 \\ NOP \\ Up: ADD W RT4 \\ Keep: NOP\end{tabular}  \\ \hline
			\end{tabular}
		}
	\end{table}
	
	\begin{tabular}[c]{@{}l@{}}  \end{tabular}
	
\section{THE DARWIN3 CHIP ARCHITECTURE}
	
\subsection{Overview}
\begin{figure*}[!t]
		\centering
		\subfloat[]{\includegraphics[scale=0.36]{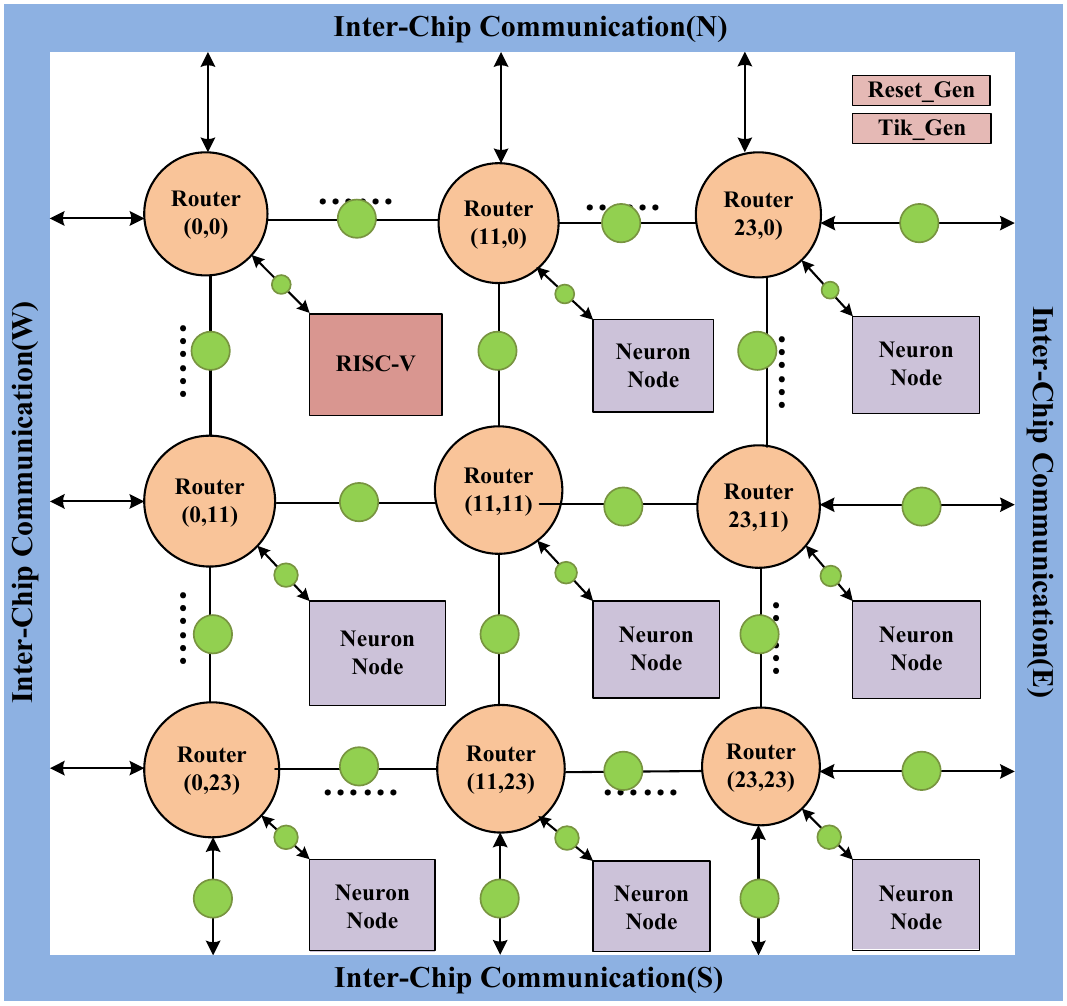}} \vspace{-4 mm}
		\subfloat[]{\includegraphics[scale=0.45]{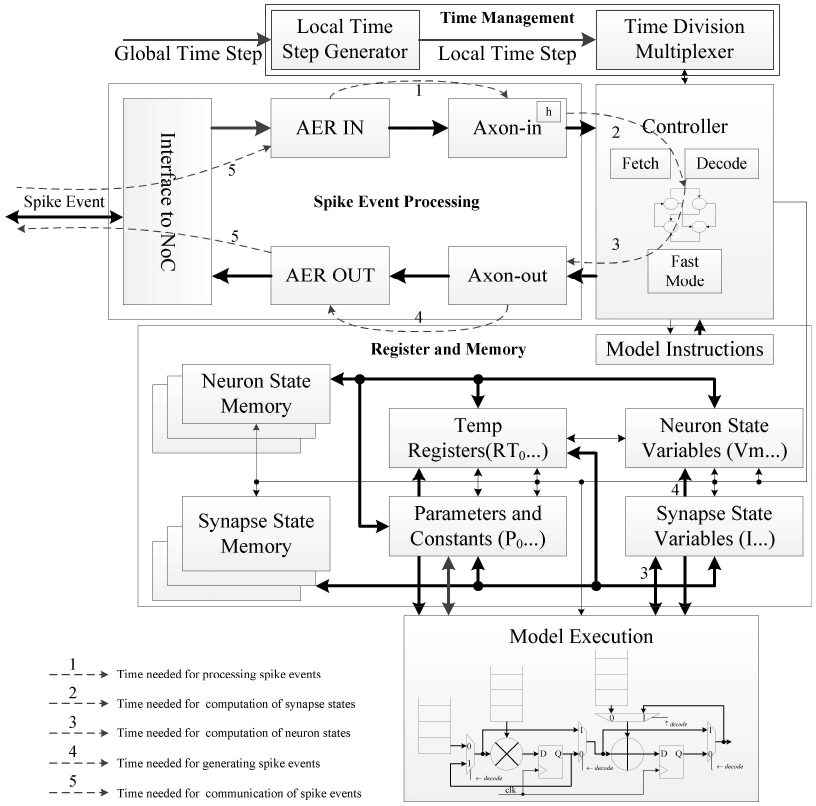} } 
		\\
		\subfloat[]{\includegraphics[scale=0.43]{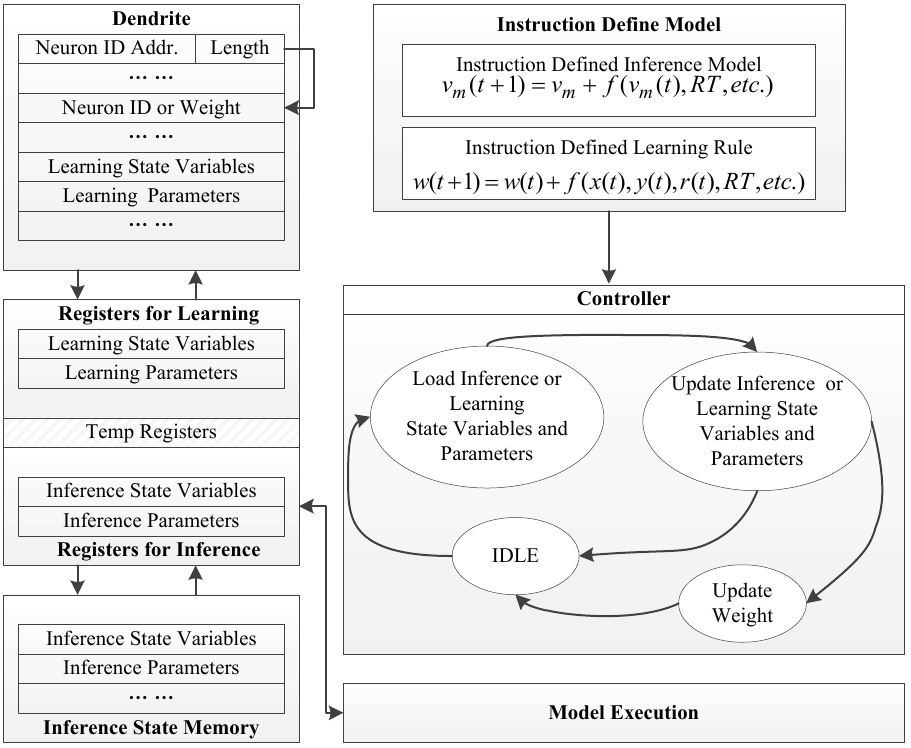}} 
		\subfloat[]{\includegraphics[scale=0.45]{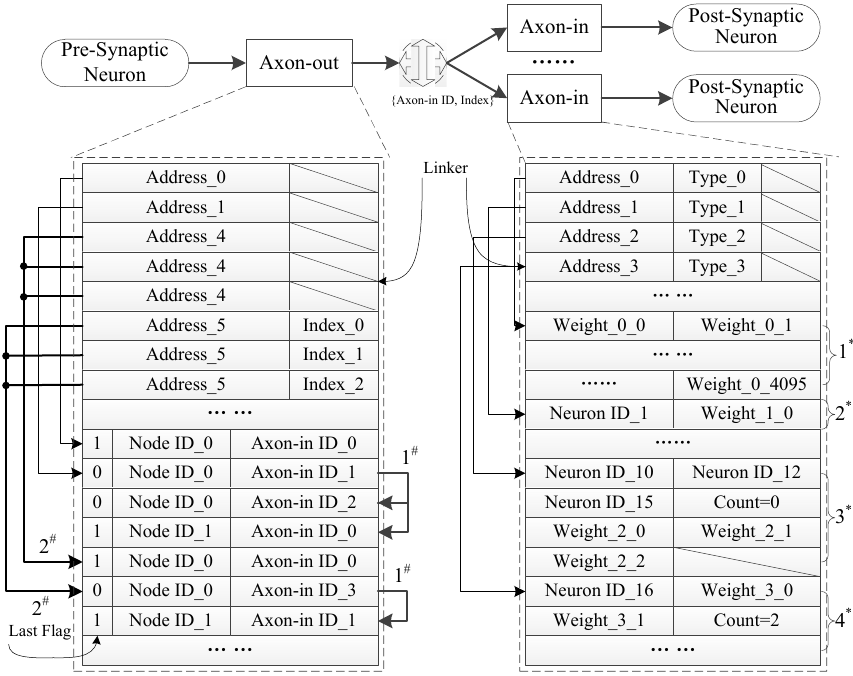}} 
		%\hfil
		%\subfloat{\includegraphics[scale=0.5]{./pics/async_sync.pdf}}
		\caption{The Architecture of The Chip Top and Main Blocks. (a) The Top Architecture of The Proposed Chip. (b) The Architecture of a Neuron Core. (c) The Architecture for Inference and Learning Process. (d) The Architecture of The Synapses.} 
		\label{fig_archeticture_all}
	\end{figure*}

The Darwin3 chip architecture is characterized by a two-dimensional mesh of computing nodes, forming a 24*24 grid, interconnected via a Network on Chip (NoC), shown in Figure \ref{fig_archeticture_all}(a). The node at position (0,0) features a RISC-V processing core for chip management, while the other nodes, functioning as neuron cores, handle the majority of computations, with each supporting up to 4096 spiking neurons. Inter-chip communication modules are placed at four edges of the chip connected with peripheral routers, acting as compression and decompression units. This design enables the NoC to extend connections to other chips in all four cardinal directions, enhancing system scalability.

This work implements a low-latency NoC architecture, which employs XY routing as detailed in \cite{NoC}. The design is further improved by integrating the CXY \cite{CXY} and OE-FAR \cite{FAR} routing strategies to tackle congestion issues. Additionally, a new routing algorithm is introduced in this work that uses the relative offsets between the source and destination addresses as the basis for its routing scheme. This strategic decision simplifies the data packet transmission process to neighboring chips, eliminating the need for complex routing protocols and address translation. 

The asynchronous communication interfaces, denoted by green circles in the figure, interconnect local synchronous modules, establishing Darwin3 as a global asynchronous local synchronous system. This enhances the capability of each node on the chip to operate independently at a high-performance level.
	
\subsection{Architecture of Neuron Cores}
\label{sec_core}
The architecture of a neuron core, illustrated in Figure \ref{fig_archeticture_all} (b), comprises five components: the controller unit, the model execution unit, the time management unit, the register and memory units, and the spike event processing unit.

The controller unit is responsible for fetching, decoding, and executing flow control. The model execution unit can perform various arithmetic and logical operations. As defined in Table \ref{regs} (a), registers store state variables, parameters, constants, and temporary variables.

The time management unit has two primary responsibilities. First, it generates an internal tick signal based on global time-step information, indicating the progression of time steps within the core. Second, it implements time-division multiplexing for 1-4096 logical neurons based on configuration information.

Each neuron core has memories for different things like axon-in, axon-out, neuron state variables, synapse state variables, and instructions. Instructions are only used to describe how neurons and synapses work. The neuron's ID determines the address of instructions and related state variables. The memories for axon-in and axon-out store how neurons are connected, and their organization is shown in Figure \ref{fig_archeticture_all} (d). When the chip starts, we need to set up the memories for the working nodes. Configuration data will be transported from the external controller (e.g., a PC or an FPGA) to the corresponding nodes through the Inter-Chip Communication module.

Unlike conventional processors that fetch instructions in every cycle using a clock, Darwin3's neuron cores are driven by spike events. When a neuron gets a spike, AER IN queries the corresponding axon-in entry to find the neuron ID and weight, calculating the state variable h. When a time step advances, the controller unit performs computations for each neuron's inference or learning stage based on the instructions. And if a neuron fires a spike, the AER OUT gets the address and ID of the post-synaptic neuron from the axon-out, packaging this into a spike data packet.

The dashed lines in Figure \ref{fig_archeticture_all} (b) illustrate the process of a neuron core receiving, processing, generating, and transmitting spikes. Multiplication operations take two cycles, while addition operations take one cycle. For example, the commonly used LIF neuron model requires four cycles (two multiplication and two addition operations), while the CUBA Delta model requires three cycles (one multiplication and one addition operation). Transmission delay is expressed as 2*N + 2*(N+1), where 2*N is for delay through N routers, and 2*(N+1) is for delay through N+1 asynchronous interconnections between pre-synaptic and post-synaptic neurons.

Figure \ref{fig_archeticture_all} (c) shows the architecture tailored for inference and learning based on the proposed ISA. In the inference mode, the controller unit updates the state variables of each neuron described by the instructions within the current time step. In the learning mode, the controller unit extracts learning parameters and state variables to execute necessary calculations and updates, calculating new weights. The axon-in memory area has been reconfigured to accommodate learning-related parameters and state variables to optimize hardware resources.
	
\subsection{Representation of Neuronal Connections}
\label{sec_synapse}
A flexible connection representation mechanism is essential in pursuing the development of neuromorphic computing chips capable of supporting complex networks. Several connection topologies find frequent application in SNNs:
	\begin{enumerate}
		\item  Multiple neuron groups connect to a group of neurons, similar to the Convolutional Neural Network (CNN) arrangement with shared weights.
		\item  A single neuron connects to an entire group of neurons.
		\item  A group of neurons fully connect to another group of neurons.
	\end{enumerate}
Upon a comprehensive examination of commonly employed connection expression mechanisms (as summarized in Table \ref{conn_mech}), we discovered that the approach used by Loihi\cite{Loihi} stands out for its exceptional flexibility, featuring substantial fan-in and fan-out capability. Combining these advantageous attributes, we have introduced a novel scheme that enables a highly compressed representation of connection topology, as depicted in Figure \ref{fig_archeticture_all} (d). To efficiently represent the topology of connections, each neuron core has independent memories for axon-out and axon-in within this framework.

Spikes are conveyed utilizing the address-event representation (AER) method. Following the generation of a spike by a pre-synaptic neuron, it accesses axon-out to retrieve the target node's address and axon-in index information of the target node. Subsequently, the AER OUT module encapsulates and transmits this information to the router through the local connection port. The router, in turn, directs the data packet towards the designated target node. Upon reception of the data packet, the target node queries axon-in to acquire pertinent information concerning the target neuron and connection weights.

Within the framework of the axon-out structure, each operational neuron is associated with a Linker. The Linker's entries retain the address of the entry containing detailed connection information and the specific index of the neuron. This index distinguishes cases in which multiple neurons are connected to the same target node. 	
This structural configuration optimizes the compression of information for connection types extending beyond point-to-point scenarios:
	
	\begin{enumerate}
		\item The last flag (LF) is set to 0 when a neuron connects to multiple nodes, indicating non-terminal nodes and facilitating efficient compression of redundant information. As shown in Figure \ref{fig_archeticture_all} (d), the situation is represented by $1{^\#}$.
		
		\item When multiple neurons connect to the same node(s), a single entry suffices for their connectivity information. As shown in Figure \ref{fig_archeticture_all} (d), the situation is represented by $2{^\#}$.
		\end{enumerate}

	\begin{table}[]
		\centering
		\caption{Different Connectivity Mechanisms}
		\label{conn_mech}
		\scalebox{0.6}{
			\begin{tabular}{lll}
				\toprule[0.5mm]
				\multicolumn{1}{c}{\textbf{Connectivity mechanism}} & \multicolumn{1}{c}{\textbf{Max. fan-in/core}} &
				\multicolumn{1}{c}{\textbf{Max. fan-out/core}} \\
				\midrule[0.25mm]
				Crossbar\cite{TrueNorth}            & C             & R  \\
				Normal index\cite{FlexLearn}        & D1            & D2                  \\
				Synaptic expansion\cite{Unicorn}    & D1            & D2*M                \\
				Population-based index\cite{Loihi}  & D1*N          & D2                  \\
				Flexible compression for Darwin3    & (D1-1)*N      & (D2-N)*N            \\
				\bottomrule[0.5mm] 
				\multicolumn{3}{p{11cm}}{ {\footnotesize $D_1$ represents fan-in memory depth (commonly associated with axon-in structures), $D_2$ represents fan-out memory depth (commonly linked to axon-out structures), M represents the number of neuron cores, and N represents the number of neurons within a neuron core. R and C represent the dimensions of the crossbar, with R being equivalent to $D_1$ (rows) and C being equivalent to $D_2$ (columns).}}    
			\end{tabular}
		}
	\end{table}
	
Each received axon-in index aligns with a corresponding Linker within the axon-in structure context. The entries in the Linker encapsulate the address of the entry, housing detailed connection information and the type of connection. This structure is strategically designed to optimize information compression, particularly tailored for connection types extending beyond point-to-point scenarios:
	\begin{enumerate}
		\item When all 4096 neurons within the node are connected to one pre-synaptic neuron, there is no necessity to store neuron indexes individually. In such cases, 4096 weights can be stored sequentially. As depicted in Figure \ref{fig_archeticture_all} (d), the case is denoted by $1{^*}$.
		\item In instances where multiple neurons are connected to a specific neuron within the node and share the same weights, storing a single neuron index along with the corresponding weight suffices. As depicted in Figure \ref{fig_archeticture_all} (d), the case is denoted by $2{^*}$.
		\item When neurons from a remote cluster are connected to a group of neurons within the target node, it is necessary to have only one instance of neuron indexes, and weights can be stored systematically based on the order of the source neurons. As depicted in Figure \ref{fig_archeticture_all} (d), the case is denoted by $3{^*}$.
        \item When the target neurons are organized sequentially, it becomes sufficient to store only the index of the initial neuron and the count of the target neurons, further reducing storage demands. As depicted in Figure \ref{fig_archeticture_all} (d), the case is denoted by $4{^*}$.
	\end{enumerate}

 This structure also facilitates the incorporation of weights with different bit widths, allowing diverse weights to be accommodated within a shared entry, consequently improving storage density.

	\begin{table*}[ht]
		\centering
		\caption{ Performance and Specifications of State-of-The-Art Neuromorphic Chips}
		\label{chip_compare}
		\scalebox{0.47}{
			\begin{tabular}{m{2.5cm}<{\centering}m{1.0cm}<{\centering}m{1.5cm}<{\centering}m{2.0cm}<{\centering}|m{1.8cm}<{\centering}m{2cm}<{\centering}m{1.4cm}<{\centering}m{1.2cm}<{\centering}m{1cm}<{\centering}m{2.1cm}<{\centering}m{0.8cm}<{\centering}m{1.6cm}<{\centering}m{1.2cm}<{\centering}m{1.4cm}<{\centering}m{2cm}<{\centering}}
				\toprule[0.5mm]
				Chip name & Neurogrid \cite{Neurogrid} & DYNAPs \cite{DYNAPs}& Brainscales2 \ \cite{BrainScaleS2}& SpiNNaker \cite{SpiNNaker}& SpiNNaker2$^{\#1}$ \cite{SpiNNaker2} & TrueNorth \cite{TrueNorth}& Loihi \cite{Loihi}& FlexLearn\cite{FlexLearn} & ISSCC 2019 \cite{isscc2019}& ODIN \cite{ODIN}& Loihi2\ \ \ \ \ \ \ \ \cite{Loihi2} & Unicorn \cite{Unicorn}&  ANP-I \ \cite{ANP}& Darwin3  \\ 
				\midrule[0.25mm]
				Implementation & Mixed & Mixed & Mixed & Digital & Digital & Digital & Digital & Digital & Digital & Digital & Digital & Digital & Digital & Digital  \\ 
				Technology(nm) & 180 & 180 & 65 & 130 & 22 & 28 & 14 & 45 & 65 & 28 & 7$^{\#2}$  & 28 & 28 & 22  \\ 
				Die area(mm2) & - &  43.79 & - & 102 & 8.76 &  430 & 60 & 410.5 & 10.08$^{\#3}$ & - & 31 & 500 & 1.628 & 358.527  \\ 
				Neuron cores & ~ & 4 & 4 & 18 & 8 & 4k & 128 & 128 & 1 & 1 & 128 & 36 & 64 & 575  \\ 
				Neurons per core & 64k & 256 & 128 & Programmable & Programmable & 256 & max. 1k & max. 35 & 400 & 256 & max. 8k & 1k & 8 & max. 4k  \\ 
				Fan in/out per core & ~ & 64/4k & 512/256 & Programmable & Programmable & 256/256 & 4k-4M/4k & - & - & - & -$^{\#4}$ & 256-256k/- & - & 28k-256M /12k-48M  \\ 
				Synaptic weight  & 4-bit & 12-bit & 6-bit & Programmable & Programmable & 1-bit & 1- to 9-bit & - & 14-bit & 3-bit & 1- to 9-bit & 4-bit & 8,10-bit & 1/2/4/8/16-bit  \\ 
				Neuron models & LIF & AdEx-IF & AdEx-IF & Programmable & Programmable & LIF & LIF & configurable & - & LIF & Programmable & LIF & LIF$^{\#5}$ & Programmable  \\ 
				Synapse models & COBA Delta & COBA NMDA & CUBA/COBA Alpha & Programmable & Programmable & CUBA Delta & CUBA Delta & configurable & CUBA \ \ \ \ \ Delta & CUBA Delta & - & CUBA Delta & CUBA Delta & Programmable  \\ 
				On-chip learning & No & No & STP/STDP/R-STDP & Programmable & Programmable & No & STDP based & configurable & Mod.SD & SDSP & Programmable & No & S-TP & Programmable  \\ 
				Energy per SOP & 941pJ @3.0V & 417fJ\cite{dynps_pow} @1.8V & - & 11.3nJ\cite{Scalable} @1.2V &  10pJ\ \ \ \ \ \ \ @0.5V & 26pJ\cite{million_truenorth} @0.775V & 23.6pJ$^{\#6}$ @0.75V & - & - & 8.4pJ @0.55V & - & - & 1.5pJ @0.56V & 5.47pJ @0.8V \\ 
				\bottomrule[0.5mm] \\
				\multicolumn{15}{p{27.6cm}}{\begin{tabular}[c]{@{}l@{}}
						\#1 Darwin3 allows nodes to operate at different frequencies, with internal modules typically running at 300-400MHz.
						\\ \#2 A test chip contains 2 QPEs with 8 PEs, while a full SpiNNker2 has 38 QPEs with 152 PEs. 
						\\ \#3 Loihi2 has been implemented in Intel 4, equivalent to the 7nm process. 
						\\ \#4 This data is obtained through a digital synthesis flow, not from the final silicon tape-out data. 
						\\ \#5 Loihi2's “Axon Routing”, which refers to fan-out or fan-in, has a topology compression of 256x.
						\\ \#6 Its output layer consists of ten integrate-and-fire(IF) neurons.  
						\\ \#7 A minimum SOP energy of 23.6 pJ at 0.75 V is extracted from pre-silicon simulations.
				\end{tabular}} 
			\end{tabular}
		}
	\end{table*}
	
	\section{EXPERIMENTAL RESULTS}
	
	\begin{figure}[!t]
		\centering
		\includegraphics[scale=0.41]{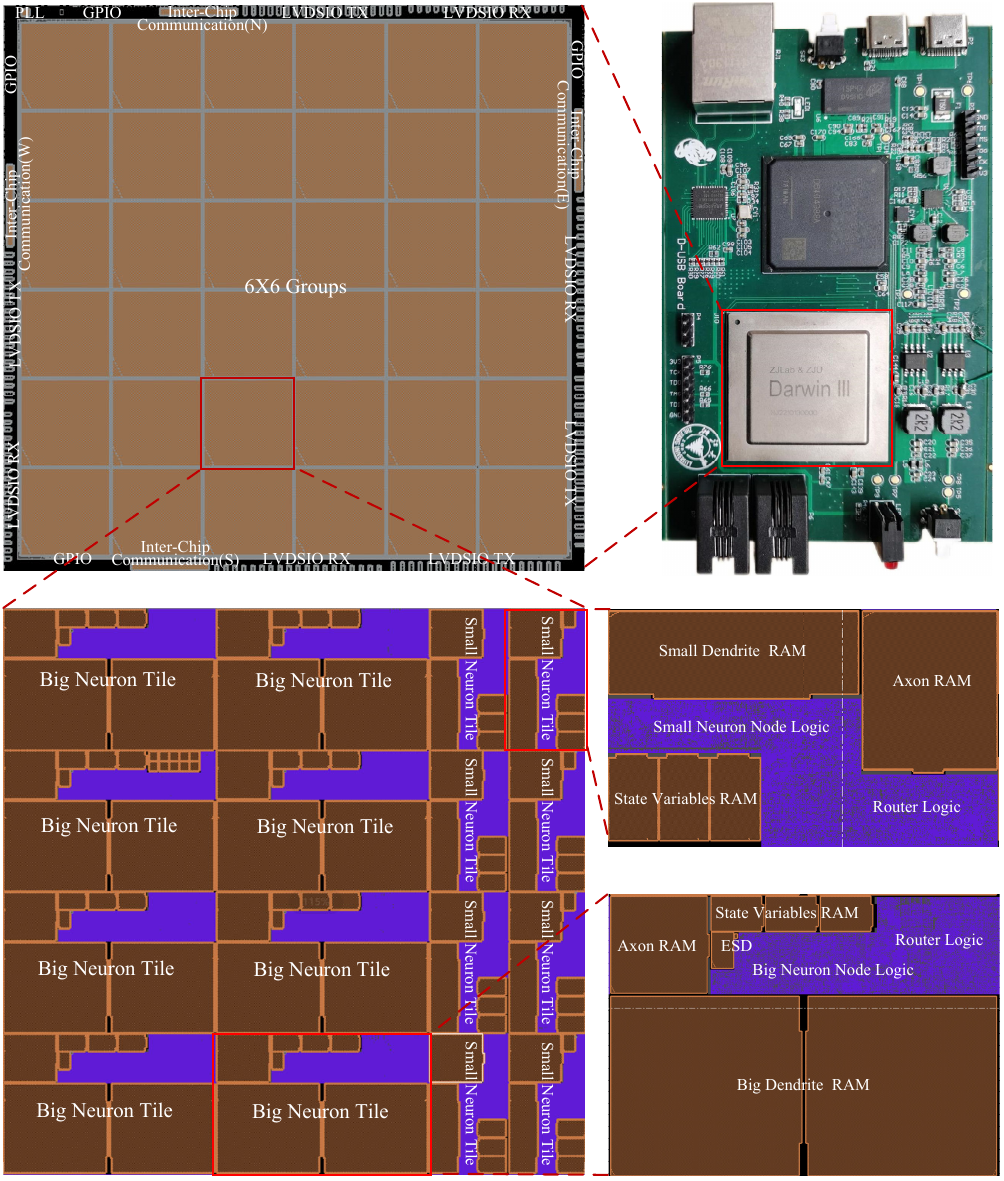}
		\caption{The Test Chip and System Board}
		\label{fig_chip}
	\end{figure}
	
To evaluate the proposed ISA and architecture, we first implemented the entire architecture in Verilog at the RTL level. Using the GLOBAL FOUNDRIES 22nm FDSOI process, we generated a GDSII file that meets the sign-off requirements after completing physical design and verification. 

After the initial Chip-on-Board(COB) testing in December 2022, the chip was repackaged using Flip-Chip BGA, and a dedicated test system board was assembled, featuring a Xilinx 7-Series FPGA. Figure \ref{fig_chip} illustrates the system board, chip layout, and main blocks. The chip's structure is organized into a grid of 6*6 groups, each consisting of 4*4 tiles. Each tile comprises a node connected to a router. Except for the RISC-V node, all nodes on the chip are neuron cores, collectively driving their computational functions. Notably, two distinct tile types exist, primarily differing in the size of their axon-in memory. We first compare some important metrics with the current state-of-the-art works, and then we run some application demonstrations to verify the chip's functionalities and performance.
	
\subsection{Comparison with The State-of-the-Art Neuromorphic Chips}	
Table \ref{chip_compare} summarizes the performance and specifications of state-of-the-art neuromorphic chips. Mixed-signal designs with analog neurons and synapse computation and high-speed digital peripherals are grouped on the left\cite{Neurogrid}\cite{DYNAPs}\cite{BrainScaleS2}, and digital designs, including Darwin3, are grouped on the right\cite{SpiNNaker}\cite{SpiNNaker2}\cite{TrueNorth}\cite{Loihi}\cite{Unicorn}\cite{FlexLearn}\cite{Loihi2}. The critical metrics for efficient spiking neuromorphic hardware platforms are the scale of neurons and synapses, model construction capabilities, synaptic plasticity, and energy per synaptic operation.
\subsubsection{Neuron Number}	
The quantity of neurons and synapses directly determines the size and complexity of the spiking neural network that a neuromorphic chip can support, which is extremely important. However, a direct comparison with the SpiNNaker chips\cite{SpiNNaker}\cite{SpiNNaker2} is not feasible due to its use of ARM processors, where the scale is tied to the size of the off-chip memory. Among other chips, NeuroGrid\cite{Neurogrid} has the largest number of neurons in a single neuron core, reaching 64K. Loihi\cite{Loihi}, Unicorn\cite{Unicorn}, Loihi2\cite{Loihi2}, and Darwin3 are at a similar level, boasting neuron counts exceeding 1K. At the chip level, Darwin3 can support up to 2.35 million neurons, surpassing the scale of TrueNorth and Loihi2 by more than two times.

\subsubsection{Synapse Capacity}
The capabilities of fan-in and fan-out within each neuron core profoundly impact the chip's overall capacity of synapses, as detailed in Table \ref{conn_mech}. Darwin3 distinguishes itself with its adaptive axon-out and axon-in memory configuration, coupled with efficient compression mechanisms, enabling remarkable fan-in and fan-out capacities of up to ($D_1$-1) * M * N and ($D_2$-N) * N$^2$, respectively. In the case of Darwin3, the compression mechanism yields a maximum fan-in improvement of 1024x and a maximum fan-out improvement of 2048x when compared to the physical memory depth.
	
	\begin{figure}[t]
		\centering
		\subfloat[]{\includegraphics[scale=0.22]{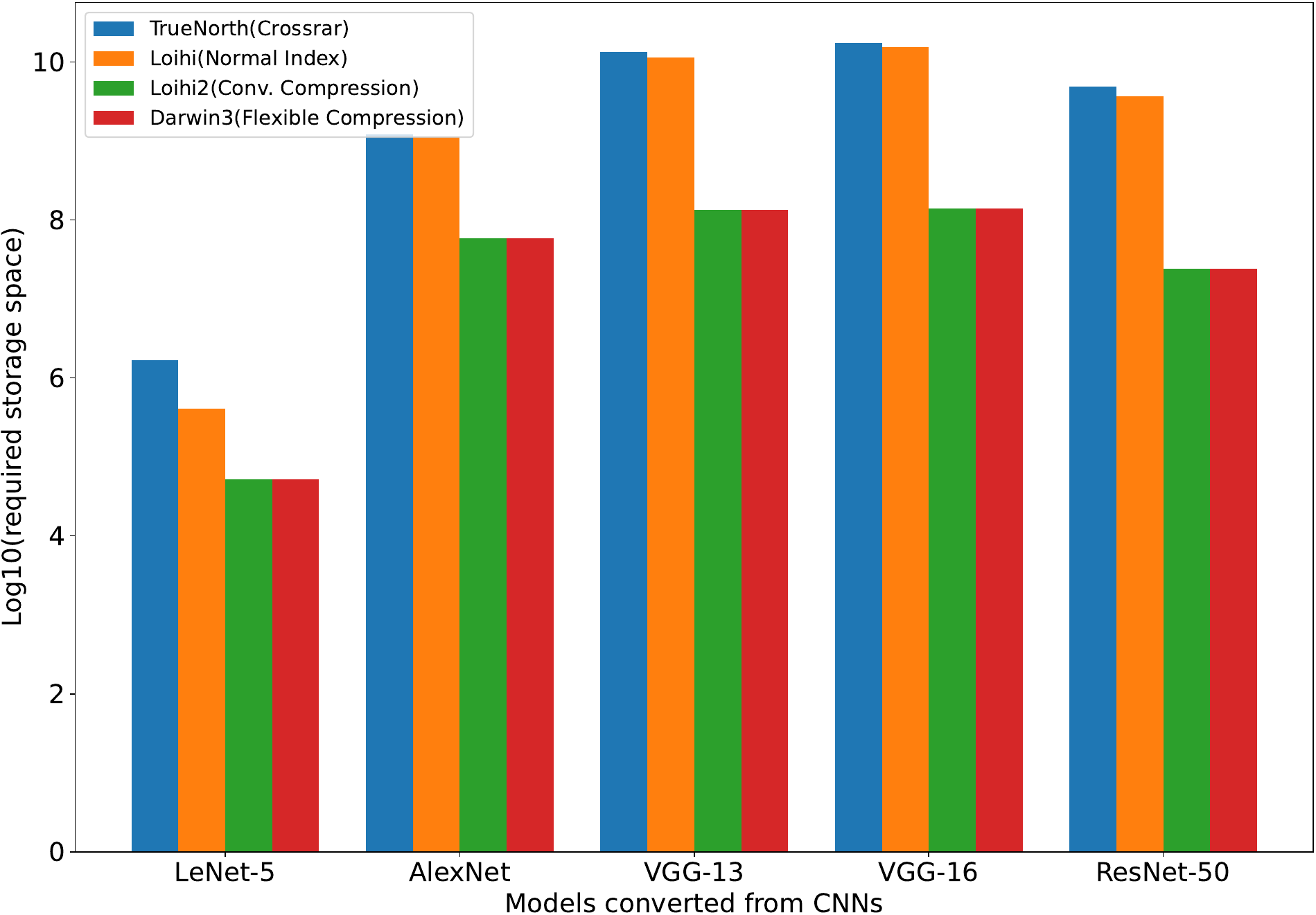}} \vspace{-4 mm}
		\hfill
		\subfloat[]{\includegraphics[scale=0.32]{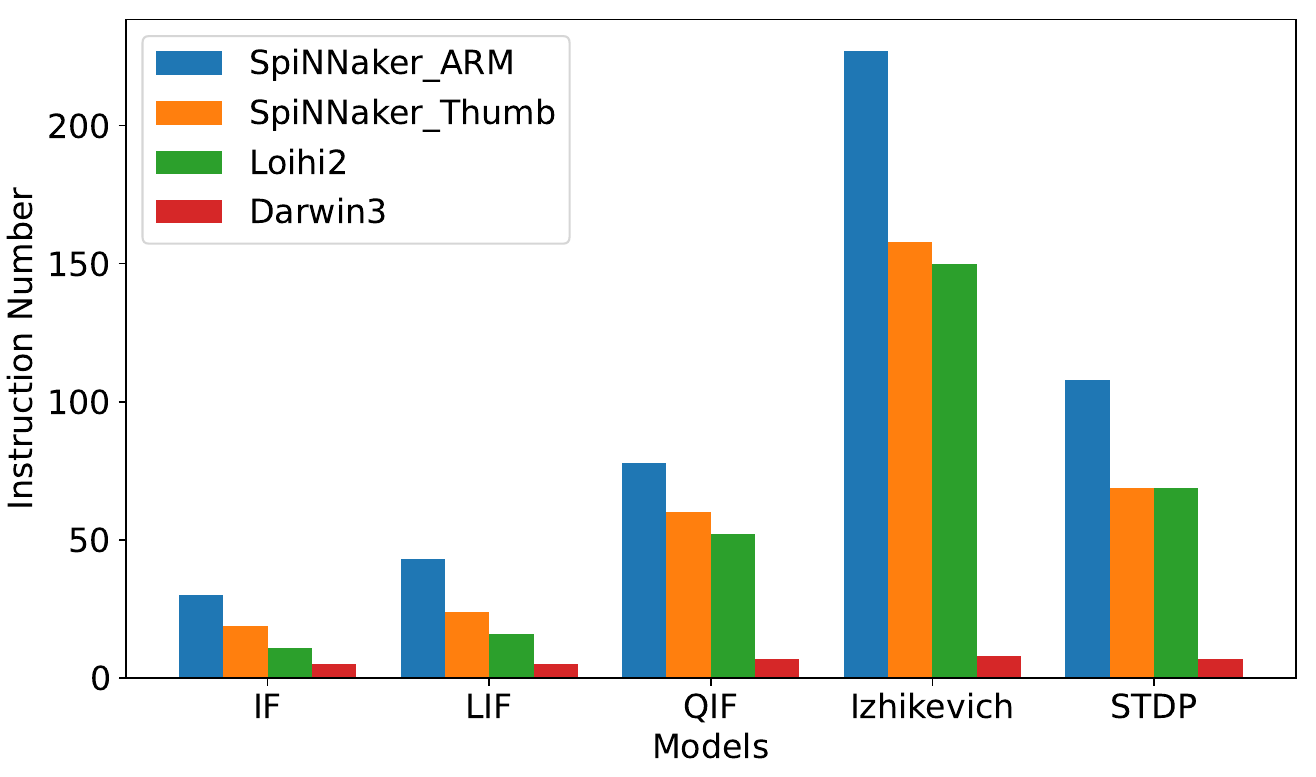} } 
		%\hfil
		\caption{Comparison of Code Density and Memory Usage. (a) Comparison of Required Weight Memory Across Typical Networks. (b) Comparison of Code Density.}
		\label{fig_comp_all}
	\end{figure}
	
 While the previous discussion delved into fan-in and fan-out capabilities, focusing on the synaptic connectivity potential, the challenge of efficiently storing synaptic weight parameters remains crucial. In Figure \ref{fig_comp_all} (a), we present a comparative analysis of weight storage requirements, highlighting the stark contrast between Darwin3 and existing approaches when applied to typical networks converted from Convolutional Neural Networks (CNNs). Chips lacking specialized compression mechanisms exhibit dense weight matrices, making memory usage 6.8 to 200 times larger than the original approach. In crossbar designs, neurons consistently occupy their unique space, contributing to additional inefficiencies.

 Darwin3 employs a versatile mechanism by classifying convolutional connections into weight-sharing multi-to-multi forms and obtains storage parity with the initial parameters, thereby achieving efficiency comparable to Loihi2\cite{Loihi2}. Importantly, this advantage extends to non-convolutional connections featuring shared weight parameters. Darwin3 enables instructional access to the complete axon-in, thus realizing the factorized attribute, which is also supported by Loihi2\cite{Loihi2}, through multiplication operations. Furthermore, Darwin3 offers compatibility with diverse weight-bit widths, enhancing its adaptability and storage efficiency.
	
	\begin{table*}[!t]
		\centering
		\caption{Performance Comparison with Other Chips}
		\label{Apps}

		(a) Inference Mode \\
\scalebox{0.5}{
	\begin{tabular}{lccccccccccccc}
		\toprule[0.5mm]
		\textbf{Platform} &
		\multicolumn{2}{c}{TrueNorth\cite{Mapping}\cite{Convolutional}} &
		\multicolumn{2}{c}{Loihi\cite{Energy}\cite{efficient} }&
		SpiNNaker\cite{Scalable}&
		ReckOn\cite{ReckOn} &
		\multicolumn{2}{c}{ANP-I\cite{ANP}} &
		\multicolumn{5}{c}{Darwin3} \\
		\midrule[0.25mm] 
		\textbf{Weight Preciency$^{\#1}$} &
		\multicolumn{1}{c}{8-bit} &
		8-bit &
		\multicolumn{1}{c}{16-bit} &
		9-bit &
		8-bit &
		8-bit &
		\multicolumn{2}{c}{8,10-bit$^{\#2}$} &
		\multicolumn{1}{c}{8-bit} &
		\multicolumn{1}{c}{16-bit} &
		\multicolumn{1}{c}{8-bit} &
		\multicolumn{1}{c}{8-bit$^{\#3}$} &
		\multicolumn{1}{c}{8-bit}\\ 
		\textbf{Frequency} &
		\multicolumn{2}{c}{-} &
		\multicolumn{2}{c}{-} &
		150 MHz &
		13MHz &
		\multicolumn{2}{c}{210MHz} &
		\multicolumn{5}{c}{333MHz} \\ 
		\textbf{DateSet} &
		\multicolumn{1}{c}{MNIST} &
		CIFAR-10 &
		\multicolumn{1}{c}{MNIST} &
		IBM Gesture &
		MNIST &
		IBM Gesture &
		\multicolumn{1}{c}{N-MNIST} &
		IBM Gesture &
		\multicolumn{1}{c}{MNIST} &
		\multicolumn{1}{c}{MNIST} &
		\multicolumn{1}{c}{MNIST} &
		\multicolumn{1}{c}{IBM Gesture} &
		\multicolumn{1}{c}{CIFAR-10} \\ 
		\textbf{Network Topology} &
		\multicolumn{1}{c}{LeNet} &
		\multicolumn{1}{c}{Mod. VGG} &
		\multicolumn{1}{c}{VGG-9} &
		cNet &
		DBN &
		RNN&
		\multicolumn{2}{c}{-} &
		
		\multicolumn{1}{c}{LeNet} &
		\multicolumn{1}{c}{VGG-9} &
		\multicolumn{1}{c}{RNN} &
		\multicolumn{1}{c}{cNet} &
		\multicolumn{1}{c}{Mod. VGG}\\
		
		\textbf{Accuracy} &
		\multicolumn{1}{c}{99.40\%} &
		83.41\% &
		\multicolumn{1}{c}{99.79\%} &
		89.64\% &
		95.01\% &
		87.30\% &
		\multicolumn{1}{c}{96.00\%} &
		92.00\% &
		\multicolumn{1}{l}{99.10\%} &
		\multicolumn{1}{c}{99.79\%} &
		\multicolumn{1}{c}{87.51\%} &
		\multicolumn{1}{c}{89.60\%} &
		\multicolumn{1}{c}{90.17\%} \\ 
		\textbf{Latency} &
		\multicolumn{1}{c}{5.74ms} &
		- &
		\multicolumn{1}{c}{6.13ms} &
		- &
		20ms &
		15ms &
		\multicolumn{2}{c}{-} &
		\multicolumn{1}{c}{5.7ms} &
		\multicolumn{1}{c}{6.48ms} &
		\multicolumn{1}{c}{2.7ms} &
		\multicolumn{1}{c}{6.08 ms} &
		\multicolumn{1}{c}{9.88ms}   \\ 
		\bottomrule[0.5mm]
		\multicolumn{12}{p{22cm}}{\footnotesize \begin{tabular}[c]{@{}l@{}}\#1 The weight precision here refers to the precision of the network run in the experiment rather than the maximum weight precision of the chip. 
				\\ \#2 The synaptic weights of its ten neurons in the output layer are 10-bit.
				\\ \#3 Darwin3 cannot support 9-bit of weight precision. \end{tabular}}
	\end{tabular}
}
		\\
		\hfill
		\\
		(b) Learning Mode \\
\scalebox{0.5}{
	\begin{tabular}{lcccccc}
		\toprule[0.5mm]
		\textbf{Platform}           & SpiNNaker\cite{Scalable} & Loihi\cite{In_hardware}   & ISSCC'2019\cite{isscc2019} & ODIN\cite{ODIN} & ANP-I\cite{ANP} & Darwin3 \\ 
		\midrule[0.25mm]
		\textbf{Frequency}          & 150 MHz   & -      & 20 MHz     & 150 MHz    & 40 MHz      & 333MHz  \\ 
		\textbf{DataSet}            & 	\multicolumn{6}{c}{MNIST} \\
		\textbf{Learning Algorithm} & CD        & EMSTDP & Mod.SD     & SDSP       & S-TP        & RSTDP   \\ 
		\textbf{Weight Preciency $^{\#1}$}   & 16-bit    & 8-bit   & 14-bit     & 3-bit      & 8-bit$^{\#2}$       & 16-bit  \\ 
		\textbf{Network Topology}   & 784-500-500-10 &   -  & (784)-200-200-10       & (256)-10 & (1024)-512-10 & (784)-100-100-10 \\ 
		\textbf{Accuracy}           & 95.01\%   & 94.70\% & 97.83\%    & 84.50\%    & 96.00\%     & 96.00\% \\ 
		\bottomrule[0.5mm]
	\end{tabular}
}
	\end{table*}	
	
\subsubsection{Code Density}	
Code density is a meaningful ISA metric, so we compare the code density of Darwin3 with the SpiNNaker chips\cite{SpiNNaker}\cite{SpiNNaker2} and Loihi2\cite{Loihi2}, the outstanding neuromorphic chips based on ISA. We use the C code to describe a model and the spinnaker tools integrated by SpyNNaker\cite{SpyNNaker} to generate assembly code for the SpiNNaker chips. Then, we compare the length of the assembly code in Figure \ref{fig_comp_all} (b). Loihi2's RISC instruction set is similar to ARM's Thumb, where spike instructions aid in curtailing spike-related instruction codes, offering a slight edge over SpiNNaker. Darwin3 shows an advantage in code density because of our proposed instructions. This instruction set concurrently loads parameters and expedites multiplication and addition with multiple parameters. Impressively, Darwin3 gets a remarkable 2.2x to 28.3x code density advantage across distinct models.
	
\subsubsection{Inference and Learning Performance}
For researchers working on SNNs, after finalizing the model, the primary focus lies on evaluating the chip's performance during application execution, with particular attention to latency and accuracy. To evaluate the capabilities of Darwin3, we conducted several experiments under two distinct scenarios: inference and learning. Table \ref{Apps} (b) compares the performance of Darwin3 to state-of-the-art neuromorphic chips in typical applications. These applications were SNNs converted from trained and quantified ANNs. We implemented the same type of network models on Darwin3, and the performance metrics indicate that Darwin3 is in the leading position regarding accuracy and latency. The accuracy is up to 6.76\%  higher, and the latency is up to 4.5x better than others. Darwin3 exhibits advantages because it has a flexible and efficient connection construction ability, which is very friendly to the converted convolutional networks. Due to the high efficiency of connection storage, it does not increase redundant spike transmission latency. The asynchronous interconnection method employed by Darwin3 has significantly reduced the communication delay between neuron cores. Darwin3 utilizes click elements\cite{click} to construct a cross-clock domain structure, enabling the completion of cross-clock domain data transfer in just two cycles. Furthermore, the related topological structures can be split and computed in parallel with more neuron cores. We attribute the observed discrepancies to the quantization operations while mapping these models to hardware, and the quantization methods employed may vary among different approaches. It's important to note that there is still room for improving latency performance by optimizing the mapping approach.
	
To further evaluate the on-chip learning capability of Darwin3, we constructed a network based on the architecture proposed by Diehl and Cook\cite{unsupervised}. We added a supervision layer, which provides positive or negative rewards based on comparing the network's output and the target during the training process, achieving the overall implementation of the RSTDP rule. The network was trained directly on Darwin3 with a weight precision of 16-bit and we achieved a classification accuracy of 96.0\% Table \ref{Apps} (a) presents the experimental results compared with prior works, demonstrating that Darwin3 is in the leading position regarding accuracy. Mod.SD algorithm is hardware-specific and performs slightly better, while Darwin3 allows flexible construction of multiple learning rules. We plan to optimize the current learning algorithm or introduce new ones to improve performance.
 	
\subsubsection{Energy Efficiency}
The energy consumption of each synaptic operation (SOP) is the most critical energy consumption metric for neuromorphic chips. We measured the energy consumption of the Darwin3 chip when running a two-layer neural network, where the neurons in the first layer can fire spikes without inputs, and the neurons in the second layer receive spikes and perform calculations. We select the common approach\cite{Scalable} to evaluate energy consumption, as detailed in Equation \ref{equ_power}.
	\begin{align}{}\label{equ_power}
		P_{total}=P_{I}+P_{B}+(P_{N}*n)+(P_{S}*s)
	\end{align}
Where $P_{I}$ is the power dissipated by a Darwin3 chip after the power-up process with no applications configured, $P_{B}$ is the baseline power, which consists of the power dissipated by all nodes enabled without running any neurons on it, $P_{N}$ is the power required to simulate a LIF neuron with a 1 ms time-step, n is the total number of neurons, $P_{S}$ is the energy consumed per synaptic event (activation of neural connections) and s are the total synaptic events. The chip operates at a frequency 333MHz with a core voltage supply of 0.8v and an IO voltage supply of 1.8v, as shown in Table \ref{chip_compare}. The measured average SOP power consumption is 5.47pj/SOP. This metric is directly influenced by factors such as the manufacturing process, power supply voltage, and operating frequency, making fair comparison challenging. However, based on the data released by prior works under typical scenarios, Darwin3 boasts a leading achievement. Darwin3's advantage lies in its internal asynchronous interconnection circuit, which enables the chip to consume very low power when there is no spike transmission or calculation. Additionally, all memories of Darwin3 will shut down during the idle phase, reducing power consumption.
	
\subsection{Applications with A Million of Neurons}	
To further illustrate the chip's efficacy, we developed two extensive applications implemented on Darwin3, spiking VGG-16 ensembling and directly-trained\cite{trained} SNN-based maze solving, shown in Figure 5. We ensembled outputs of five VGG-16 models obtained through ANN2SNN\cite{fast} using a voting mechanism\cite{volting}, culminating in a composite model comprising approximately 1.05 million neurons and employing 8-bit weight precision. We applied random transformations to the input and used five independent VGGs in the hidden layers for the classification tasks. The voting layer produces the final classification outcome based on the collective votes from the individual outputs of the hidden layers. Compared to the original single VGG-16, accuracy testing on the CIFAR-10 dataset witnessed an increase from 92.98\% to 93.48\%.

We also developed an application for maze solving. We mapped the maze onto a set of neurons, where excitatory neurons represent the free-walking grid points, and inhibitory neurons represent obstacles. The interconnected excitatory neurons can transmit spikes in sequence, and under the action of STDP rules, the synaptic weights are continuously increased to form a stable synaptic strength. However, synapses connected to inhibitory neurons cannot be strengthened, and the transmission of spikes will be terminated when encountering inhibitory neurons. After learning, the model can quickly find the path by observing the path along which the spikes propagate. We conducted experiments using mazes of different sizes, comparing the time it takes to search for a path on our chip versus a CPU server. A map of size 15434*1534 requires over 2.35 million neurons, approaching the upper limit of neurons that Darwin3 can simulate. The result is shown in Table \ref{table_maze}. With the STDP-based SNN method, the time consumed increases linearly with the maze size. In contrast, the traditional search method on the CPU  server consumes a lot of time because it relies on many recursive operations.
	\begin{table}[b]
		\caption{Time Cost for Maze Solving Application}
		\label{table_maze}
		\scalebox{0.63}{
			\begin{tabular}{|m{2.6cm}<{\centering}|m{4.0cm}<{\centering}|m{2.8cm}<{\centering}|}
				\hline
				Maze Size (Neuron \#) & Time of Intel Xeon Gold 6248R @ 3GHz, 205W (ms) & Time of Darwin3 @ 400Mz, 1.8W (ms) \\ \hline
				63*63     & 83                         & 128          \\ \hline
				125*125   & 296                        & 412          \\ \hline
				250*250   & 1132                       & 1375         \\ \hline
				500*500   & 4744                       & 4200         \\ \hline
				600*600   & 7968                       & 6248         \\ \hline
				700*700   & 11283                      & 7706         \\ \hline
				900*900   & FAIL                       & 9654         \\ \hline
				%1000*1000 & FAIL                       & 12600        \\ \hline
				1534*1534 & FAIL                       & 22089        \\ \hline
				\multicolumn{3}{p{10cm}}{\footnotesize The mazes are randomly generated, and the running time is an average of five measurements.}
			\end{tabular}
		}
	\end{table}
	\begin{figure}[t]
		\centering
		\subfloat[]{\includegraphics[scale=0.52]{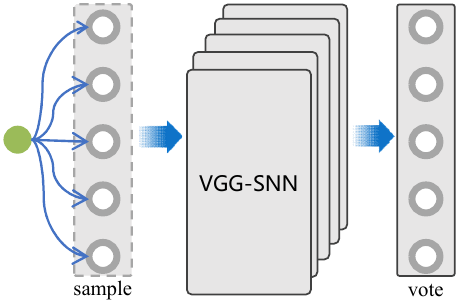}\label{fig_maze_1}} \vspace{-4 mm}
		\subfloat[]{\includegraphics[scale=0.52]{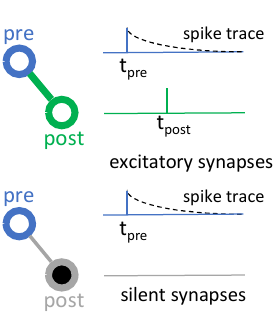}\label{fig_maze_2}} 
		\hfil
		\subfloat[]{\includegraphics[scale=0.52]{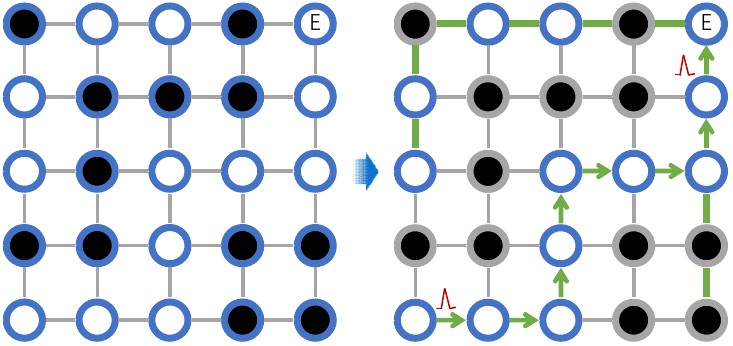}\label{fig_maze_3}}
		\caption{Two Large-scale Applications with A Million of Neurons. (a) Spiking VGG-16 Ensembling. (b) Directly-trained SNN-based Maze Solving.} 
		\label{fig_maze}
	\end{figure}

\section{CONCLUSION}
The article proposes a new instruction set and a connectivity compression mechanism to create a chip that can support large-scale neural networks. This chip has been designed to be more efficient in terms of the number of neurons it can accommodate and its synaptic computing performance, compared to existing works. The experimental results show that the chip has reached the same leading level as the state-of-the-art works in terms of accuracy and latency performance metrics, both for inference and learning modes. The practical effectiveness of the chip has also been demonstrated by running a maze-searching application on it.

Due to the chip's versatile chip communication mechanism, different Darwin3 chips can be integrated onto a single board and interconnect several boards to configure a big chassis. These chassis can be interconnected through a network infrastructure to support the construction of extensive SNNs. This configuration can support the construction of extensive SNNs when coupled with suitable software frameworks. 

%\section{SUPPLEMENTARY DATA}

\section{FUNDING}
This work is supported by the National Key R\&D Plan of China (No. 2022YFB4500100), the National Natural Science Foundation of China (No. 61925603, U20A20220, 62076084), the Key Research Project of Zhejiang Lab (No. 2021KC0AC01), the Key Research and Development Program of Zhejiang Province in China (No. 2020C03004), the grants from Key R\&D Program of Zhejiang (No. 2022C01048).
	
\section{AUTHOR CONTRIBUTIONS}
G.P. and D. M. proposed the idea. D. M., X. J., S. S., and Y. L. proposed the chip's architecture and implemented the design. X. J., S. S., Y. H., F. Y., H. T., P. L., and X. Z. built the experimental system. D. M., X. J., and S. S. completed the experiment. D. M., X. J., and X. W. wrote the article. All authors analyzed the data, discussed the results, and commented on the manuscript.
%%%%%%% -- article CONTENT ENDS -- %%%%%%%%

%%%%%%%%% -- BIB STYLE AND FILE -- %%%%%%%%
\bibliographystyle{unsrt}
\bibliography{refs}
	
\end{document}